\documentclass[letterpaper]{article} 
\usepackage{aaai25}  
\usepackage{times}  
\usepackage{helvet}  
\usepackage{courier}  
\usepackage{multirow}
\usepackage{booktabs}
\usepackage{tcolorbox}
\tcbuselibrary{listingsutf8}
\usepackage[hyphens]{url}  
\usepackage{graphicx} 
\usepackage{natbib}  
\usepackage{caption} 
\usepackage{algorithm}
\usepackage{algorithmic}

\usepackage{newfloat}
\usepackage{listings}
\DeclareCaptionStyle{ruled}{labelfont=normalfont,labelsep=colon,strut=off} 
\floatstyle{ruled}
\newfloat{listing}{tb}{lst}{}
\floatname{listing}{Listing}
\title{Mason NLP-GRP at \#SMM4H-HeaRD 2025: Prompting Large Language Models to Detect Dementia Family Caregivers}
\author {
    Md Badsha Biswas\textsuperscript{\rm 1},
    Özlem Uzuner\textsuperscript{\rm 2}
}
\affiliations {
    \textsuperscript{\rm 1}Department of Computer Science, George Mason University, Fairfax, VA, USA\\
    \textsuperscript{\rm 2}Department of Information Sciences \& Technology, George Mason University, Fairfax, VA, USA\\
    mbiswas2@gmu.com, ouzuner@gmu.edu,
}

\begin{document}

\maketitle

\begin{abstract}
Social media, such as Twitter, provide opportunities for caregivers of dementia patients to share their experiences and seek support for a variety of reasons. Availability of this information online also paves the way for the development of internet-based interventions in their support. However, for this purpose, tweets written by caregivers of dementia patients must first be identified. This paper demonstrates our system for the SMM4H 2025 shared task 3, which focuses on detecting tweets posted by individuals who have a family member with dementia. The task is outlined as a binary classification problem, differentiating between tweets that mention dementia in the context of a family member and those that do not. Our solution to this problem explores large language models (LLMs) with various prompting methods. Our results show that a simple zero-shot prompt on a fine-tuned model yielded the best results. Our final system achieved a macro F1-score of 0.95 on the validation set and the test set. Our full code is available on GitHub \footnote{https://github.com/badshabiswas/smm4h-task-3}

\end{abstract}

\section{Introduction}

Dementia is a long-term condition caused by brain disease or injury, leading to the loss of two or more cognitive abilities \cite{arvanitakis2019diagnosis}. Around 47 million people worldwide have dementia, and this number may increase to 131 million by 2050 \cite{prince2015world}.

The 10th Social Media Mining for Health (SMM4H) \cite{smm4h-heard-overview-2025} workshop introduced six shared tasks. Task 3 focused on Detection of Dementia of Family Caregivers on Twitter. The goal of this task is the development of systems that distinguish between tweets in which the author shares that a family member has dementia and those that merely mention the term without personal relevance. As a benchmark, BERTweet\cite{nguyen2020bertweet} achieves an F1 score of 0.96 \cite{info:doi/10.2196/39547} on this task. In this system description paper, we present our experiments using different prompting strategies and large language models (LLMs) and compare their performance against the benchmark. 

\section{System Description}
Our approach followed three main steps: (1) data preparation through oversampling\cite{kim2023optimal}, (2) prompt design for effective classification, and (3) model fine-tuning using LoRA (Low-Rank Adaptation). We applied these approaches on two open source models: Llama-3.1-8B-Instruct \cite{patterson2022carbonfootprintmachinelearning} and Mistral-7B-Instruct-v0.3 \cite{jiang2023mistral7b}, chosen for their size and recent superior performance on standard evaluations (MMLU and AGIEval).

\subsection{Dataset Preparation}
The dataset consists of 8846 tweets and the distribution is given in Table \ref{table1}.
\begin{table}[h]
\centering

\begin{tabular}{@{}cccc@{}}
\toprule
\multirow{2}{*}{Dataset} & \multicolumn{2}{c}{Label} & \multirow{2}{*}{Total} \\ \cmidrule(lr){2-3}
                         & 1           & 0           &                        \\ \midrule
Training                 & 4523        & 2201        & 6724                   \\ \midrule
Validation               & 234         & 119         & 353                    \\ \bottomrule
\end{tabular}
\caption{Data Distribution for Task 3}
\label{table1}
\end{table}
Each tweet is labeled 1 (the tweet’s author has a family member with dementia) or 0 (the tweet merely mentions dementia ). We initially started with various data preprocessing techniques(e.g., hashtag segmentation, stop word removal, and stemming, etc. ). However, we discarded these steps due to the negative impact on performance and proceeded with the raw data, as it contained significant and contextually important information essential for this particular task.

 The dataset is highly imbalanced, where approximately one-third of the training tweets are labeled 0, while two-thirds are labeled 1. This imbalance (roughly 33\% vs 67\%) can bias a model towards always predicting the majority class (label 1). To address this issue, for each epoch, we resampled the training examples such that label 0 tweets were replicated roughly twice so that the model saw an approximately balanced number of 0 and 1 examples. We found this approach more straightforward than adjusting loss weights.


\begin{table*}[t]
\centering

\begin{tabular}{@{}cccccccc@{}}
\toprule
\multirow{2}{*}{Prompt} & \multicolumn{3}{c}{Llama 3.1- 8B-Instruction} & \multicolumn{3}{c}{Mistral 7B-Instruction} \\ \cmidrule(l){2-7} 
                           & Precision   & Recall   & F1        & Precision   & Recall  & F1     \\ \midrule
Zero-Shot                   & \textbf{0.919}       & \textbf{0.974}    & \textbf{0.946}       & \textbf{0.956}       & 0.932   & \textbf{0.944}  \\
Few-Shot                  & 0.897       & 0.855    & 0.875           & 0.721            &  \textbf{0.996}       &  0.837      \\
Chain of Thought (COT)             & 0.891       & 0.838    & 0.863            & 0.828            & 0.966        &  0.892      \\
Cascade                    & 0.851       & 0.927    & 0.888            & 0.866             &   0.868      &   0.867    \\ \bottomrule
\end{tabular}
\caption{Performance on Validation set of LLaMA and Mistral models using different prompt strategies on Task 3}
\label{table2}
\end{table*}

\subsection{Prompt Designing}

We designed prompts using various prompt engineering strategies and then converted the original dataset into instruction-based formats following a template structure. These instructions were used to fine-tune the LLMs via LoRA adapters \cite{hu2021loralowrankadaptationlarge}. To leverage the instruction-following capabilities of the LLM, we structured the classification task as a conversational prompt. Each training instance was formatted as a brief dialogue using special tokens and role indicators.
\begin{figure}[h]
\centering
\includegraphics[width=0.9\columnwidth]{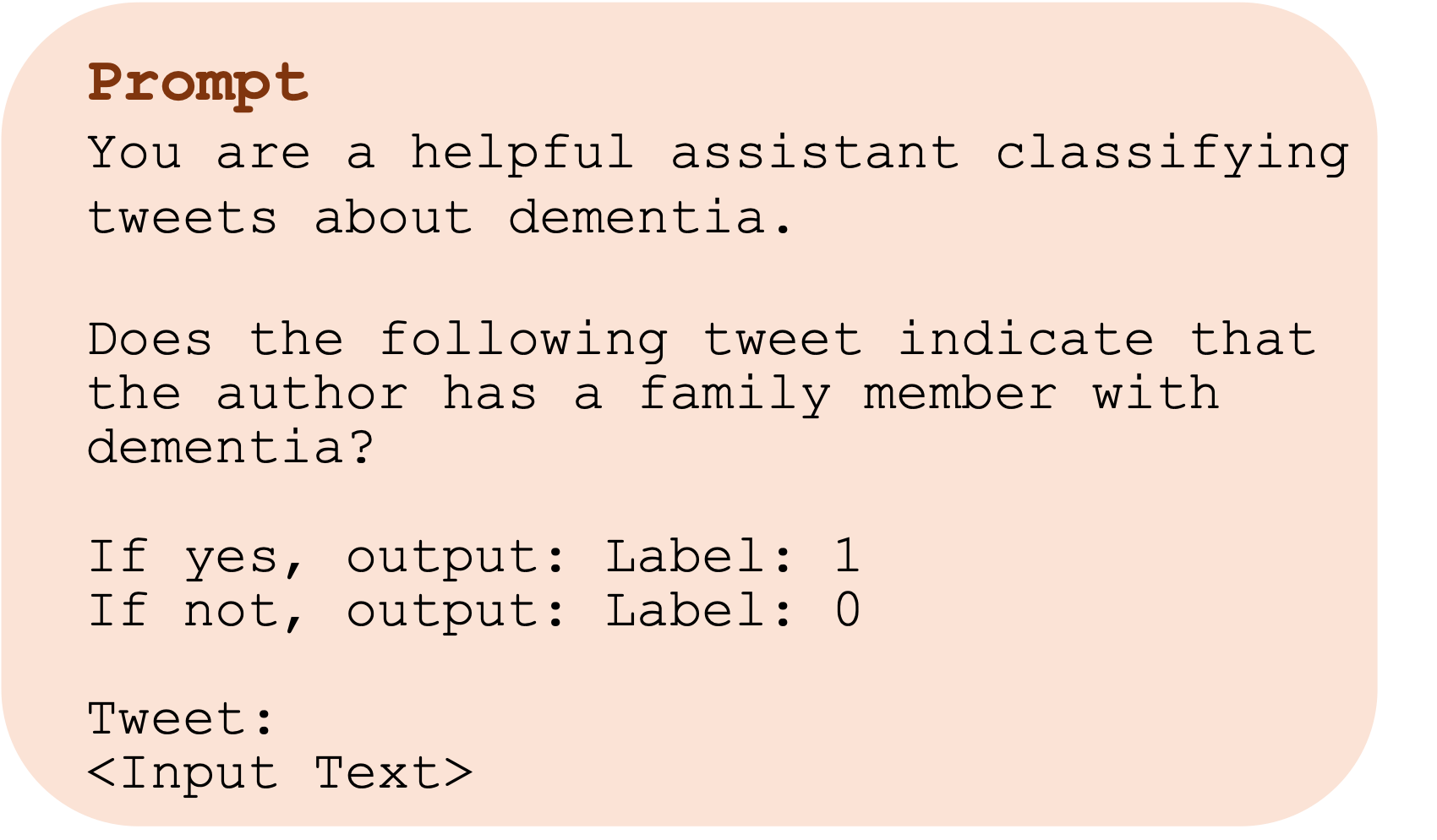} 
\caption{Illustration of prompt }
\label{fig1}
\end{figure}

We experimented with multiple prompting approaches, including zero-shot \cite{zhou2023largelanguagemodelshumanlevel}, few-shot \cite{brown2020languagemodelsfewshotlearners}, chain-of-thought \cite{wei2023chainofthoughtpromptingelicitsreasoning}, and cascaded prompting \cite{wu2024cascade} (see Appendix). Among these, our best-performing prompt, which is zero-shot, is shown in Figure \ref{fig1}.



\subsection{Model Fine-tuning}
Since fine-tuning a large language model is resource-intensive, we employed Low-Rank Adaptation (LoRA)\cite{hu2021loralowrankadaptationlarge}. We applied it to both Llama-3.1-8B-Instruct and Mistral-7B-Instruct-v0.3.
\subsubsection{Low-Rank Adaptation (LoRA)} allows efficient fine-tuning by injecting trainable low-rank weight matrices into the model's layers instead of updating all the parameters. We further combined LoRA with 4-bit quantization following the QLoRA \cite{dettmers2023qloraefficientfinetuningquantized} approach for the reduction of GPU memory usage.
\subsubsection{Instruction Tuning}
The significance of prompt templates has been highlighted in several information extraction studies, especially in the context of LLMs \cite{lu-etal-2021-text2event, bao2023opiniontreeparsingaspectbased}. This format leverages the model’s instruction-tuning to follow the given task definition. During fine-tuning, the model learns to produce the correct label given the tweet and the instruction context.
We fine-tuned the model using the HuggingFace Transformers\cite{wolf2020huggingfacestransformersstateoftheartnatural} and TRL (Transformer Reinforcement Learning) library’s SFTTrainer. The training was run for 5 epochs on the oversampled training set. We used a constant learning rate (2e-4) throughout training (no decay schedule) to keep the fine-tuning stable; 
We optimize the model parameters using the AdamW optimizer~\cite{loshchilov2019decoupledweightdecayregularization}. All experiments are conducted on an NVIDIA A100.80gb GPU.

\section{Results}
We conducted extensive experiments with various prompting techniques, as our primary goal was to compare the performance of different LLM prompting strategies. Consequently, we did not focus on optimizing the overall classification score. The evaluation metric for this task is the macro F1-score across the binary classes. Table \ref{table2} summarizes the performance of the LLaMA and Mistral fine-tuned models under various prompting techniques on the validation set. The final result on the test set is given in Table \ref{table3}

\begin{table}[h]
\centering

\begin{tabular}{@{}cccc@{}}
\toprule
               & F1-Score & Precision & Recall \\ \cmidrule(l){2-4}
Baseline\textsubscript{BERTweet}       & 0.962    & 0.946     & 0.979  \\
Our Submission & \textit{0.954}    &  \textit{0.946}     &  \textit{0.962}  \\
Mean           & 0.885    & 0.925     & 0.892  \\
Median         & 0.953    & 0.946     & 0.969  \\ \bottomrule
\end{tabular}
\caption{Task 3 official results on the testing subset}

\label{table3}
\end{table}

We also explored various training strategies, including settings with and without oversampling, with and without data preprocessing, and using different learning rates and numbers of training epochs. In the end, a learning rate of 2e-4 and 5 epochs gave us the best performance.


\section{Conclusion}
We presented a system for SMM4H 2025 Task 3 using instruction-tuned LLMs (LLaMA-3.1–8B and Mistral 7B) fine-tuned with LoRA to classify dementia caregiver tweets. Our results show that prompt-based, parameter-efficient tuning achieves strong and competitive performance to specialized tasks with some limitations (Appendix A.5). We believe that using larger models (e.g., 70B) could further improve performance and plan to explore this in future work, as this study focused primarily on evaluating the impact of prompting strategies.
\section{Acknowledgments}
The experiments were run on HOPPER clusters provided by the Office of Research Computing at George Mason University.

\bibliography{aaai25}

\section{Appendix}

\subsection{A.1 Variation of Zero-Shot Prompting}




\begin{tcolorbox}[colback=gray!5, colframe=gray!80, title=A.2 Zero-Shot Prompt Variation]
\textbf{System:} You are a classifier. Determine if the following tweet implies the author has a family member with dementia.\\

\textbf{Instruction:} Reply with '1' if the tweet indicates the author has a family member (e.g., parent, grandparent, spouse, sibling) with dementia/Alzheimer's. Reply with '0' if not (e.g., the tweet only talks about dementia in general, about someone else's family, or the author themselves).\\

Respond with a single character: 0 or 1.\\

\textbf{Tweet:}
\textless Input Text\textgreater\\

\textbf{Label} : \textless Label\textgreater
\end{tcolorbox}






\subsection{A.2 Chain of Thought Prompting}
\begin{tcolorbox}[colback=gray!5, colframe=gray!80, title=A.4 Chain of Thought Prompt]
\textbf{System:} You are a classifier. Analyze the tweet and decide if it indicates the author has a family member with dementia.\\

\textbf{Tweet:} 
\textless Input Text\textgreater\\

Let's think step by step:\\

1. Identify relevant details in the tweet (personal pronouns, family terms, etc).\\

2. Determine if the author is talking about their own family member with dementia.\\

3. If yes, the label is 1. If not, the label is 0.\\

Reasoning:\\

Conclusion:\\

\textbf{Label} : \textless Label\textgreater
\end{tcolorbox}

\subsection{A.3 Few-Shot Prompting}
\begin{tcolorbox}[colback=gray!5, colframe=gray!80, title=A.5 Few-Shot Prompt]
\textbf{System:} You are a classifier. Determine if the following tweet implies the author has a family member with dementia.\\

Example 1: \\

Tweet: "My mom has dementia and doesn't recognize me sometimes." \\
Label: 1\\

Example 2: \\

Tweet: "Dementia is such a cruel disease. Watching the news about it is heartbreaking." \\
Label: 0\\

Example 3: \\

Tweet: "My friend’s grandmother has Alzheimer’s; it’s so sad to see." \\
Label: 0\\

Now, classify the following tweet:\\

\textbf{Tweet:} 
\textless Input Text\textgreater
\\

\textbf{Label} : \textless Label\textgreater

\end{tcolorbox}

\subsection{A.4 Cascade Prompting}

\begin{tcolorbox}[colback=gray!5, colframe=gray!80, title=A.6 Cascaded Prompt Example (Training Format)]
\textbf{System:} You are a helpful assistant. You classify tweets about dementia/Alzheimer's in two steps.

Step 1: Determine if dementia/Alzheimer's is mentioned. 

Step 2: If yes, determine if a family member has dementia/Alzheimer's. Then produce a final label: 1 or 0.\\

\textbf{User:} Step 1: Does the following tweet mention dementia/Alzheimer's? Answer "Yes" or "No".\\
Tweet: \textless Input Text\textgreater\\
\textbf{Assistant:} Yes\\

\textbf{User:} Step 2: If Step 1 is "Yes", does the tweet indicate a family member with dementia/Alzheimer's? Answer "Yes" or "No".\\
\textbf{Assistant:} Yes\\

\textbf{User:} Finally, produce the classification label (1 = has family member with dementia/Alzheimer's, 0 = otherwise).\\
\textbf{Assistant:} \textless Label\textgreater

\end{tcolorbox}

\subsection{A.5 Error Analysis}
We observed the tweets that were misclassified by the model. We found that most of the errors were false positives-cases where the model predicted the presence of a family member with dementia, but the tweet did not clearly indicate a family member with dementia. For example, in the tweet "Dementia and OCDs run in the family. I aint afraid to admit I got issues that messes with my emotions. My sister too. Well guess who got the dementia." the mention of dementia and family might have misled the model, given that it is ambiguous who in the family is really affected. Another example is "@CoffeeAndKink My siblings know and are befuddled at best, concern troll- y at worst. My step dad doesn't know because there's no need for him to know. He's aging, undiagnosed but probably has dementia, and I can't see what purpose sharing the information could possibly serve." This discusses a family member but lacks a confirmed diagnosis, making the relationship unclear. These examples show that the model sometimes overgeneralizes from keywords like "family" and "dementia". On the other hand, it fails to interpret subtle or implicit relationships. For instance, in the tweet "fellas, is it gender affirmation if it's your mom's 80 year old dad with dementia who forgot you were visiting and yelled, "who's that young man breaking in??" to your mom when he saw you enter the house?" the author refers to their grandfather having dementia. While the condition is mentioned, the relationship (grandparent) is only indirectly conveyed through layered phrasing. The model failed to link the mention of dementia to a family member, resulting in a false negative. A larger model, with better coreference resolution, is more likely to recognize implicit relationships and correctly detect a family member with dementia.

\end{document}